\documentclass[10pt,conference,a4paper]{IEEEtran}
\usepackage{cite}
\usepackage{amsmath}
\usepackage{algorithmic}
\usepackage{array}
\usepackage{fixltx2e}
\usepackage{stfloats}
\usepackage{multirow}
\usepackage{booktabs}
\usepackage[caption=false,font=normalsize,labelfont=sf,textfont=sf]{subfig}
\usepackage[pdftex]{graphicx}
\graphicspath{{Image/}}

\ifCLASSINFOpdf
\else
\fi
\begin{document}
%
\title{Learning Topics using Semantic Locality}

\author{\IEEEauthorblockN{Ziyi Zhao, Krittaphat Pugdeethosapol, Sheng Lin, Zhe Li, Caiwen Ding, Yanzhi Wang, Qinru Qiu}
\IEEEauthorblockA{Department of Electrical Engineering \& Computer Science\\
Syracuse University\\
Syracuse, NY 13244, USA \\
\texttt{\{zzhao37, kpugdeet, shlin, zli89, cading, ywang393, qiqiu\}@syr.edu}}
}


%


\maketitle

\begin{abstract}
The topic modeling discovers the latent topic probability of the given text documents.
To generate the more meaningful topic that better represents the given document, we proposed a new feature extraction technique which can be used in the data preprocessing stage. The method consists of three steps. First, it generates the word/word-pair from every single document. Second, it applies a two-way TF-IDF algorithm to word/word-pair for semantic filtering. Third, it uses the K-means algorithm to merge the word pairs that have the similar semantic meaning.

Experiments are carried out on the Open Movie Database (OMDb), Reuters Dataset and 20NewsGroup Dataset. The mean Average Precision score is used as the evaluation metric. Comparing our results with other 
state-of-the-art topic models, such as Latent Dirichlet allocation and traditional Restricted Boltzmann Machines. Our proposed data 
preprocessing can improve the generated topic accuracy by up to 12.99\%.
\end{abstract}


%
\IEEEpeerreviewmaketitle

\section{Introduction}
During the last decades, most collective information has been digitized to form an immense database distributed across the Internet.
Among all, text-based knowledge is dominant because of its vast availability and numerous forms of existence. For example, news, articles, or even Twitter posts are various kinds of text documents. On one hand, it is difficult for human users to locate one's searching target in the sea of countless texts without a well-defined computational model to organize the information.
On the other hand, in this big data era, the e-commerce industry takes huge advantages of machine learning techniques to discover customers' preference.
For example, notifying a customer of the release of ``Star Wars: The Last Jedi'' if he/she has ever purchased the tickets for ``Star Trek Beyond'';
recommending a reader ``A Brief History of Time'' from Stephen Hawking in case there is a ``Relativity: The Special and General Theory'' from Albert Einstein in the shopping cart on Amazon.
The content based recommendation is achieved by analyzing the theme of the items extracted from its text description.

Topic modeling is a collection of algorithms that aim to discover and annotate large archives of documents with thematic information\cite{blei2012probabilistic}. 
Usually, general topic modeling algorithms do not require any prior annotations or labeling of the document while the abstraction is the output of the algorithms.
Topic modeling enables us to convert a collection of large documents into a set of topic vectors. Each entry in this concise representation is a probability of the latent topic distribution. By comparing the topic distributions, we can easily calculate the similarity between two different documents\cite{steyvers2007probabilistic}. The availability of many manually categorized online documents, such as Internet Movie Database (IMDb) movie review \cite{IMDB}, Wikipedia articles, makes the testing and validation of topic models possible. 

Some topic modeling algorithms are highly frequently used in text-mining\cite{mei2008topic}, preference recommendation\cite{wang2011collaborative} and computer vision\cite{wang2008spatial}. Many of the traditional topic models focus on latent semantic analysis with unsupervised learning \cite{blei2012probabilistic}. Latent Semantic Indexing (LSI) \cite{landauer2006latent} applies Singular-Value Decomposition (SVD) \cite{golub1970singular} to transform the term-document matrix to a lower dimension where semantically similar terms are merged. It can be used to report the semantic distance between two documents, however, it does not explicitly provide the topic information. The Probabilistic Latent Semantic Analysis (PLSA)\cite{hofmann1999probabilistic} model uses maximum likelihood estimation to extract latent topics and topic word distribution, while the Latent Dirichlet Allocation (LDA) \cite{blei2003latent} model performs iterative sampling and characterization to search for the same information. Restricted Boltzmann Machine (RBM) \cite{hinton2009replicated} is also a very popular model for the topic modeling. By training a two layer model, the RBM can learn to extract the latent topics in an unsupervised way.

All of the existing works are based on the bag-of-words model, where a document is considered as a collection of words. The semantic information of words and interaction among objects are assumed to be unknown during the model construction. Such simple representation can be improved by recent research in natural language processing and word embedding. In this paper, we will explore the existing knowledge and build a topic model using explicit semantic analysis.

This work studies effective data processing and feature extraction for topic modeling and information retrieval. We investigate how the available semantic knowledge, which can be obtained from language analysis, can assist in the topic modeling. 

Our main contributions are summarized as the following:
\begin{itemize}
  \item A new topic model is designed which combines two classes of text features as the model input.
  \item We demonstrate that a feature selection based on semantically related word pairs provides richer information thank simple bag-of-words approach. 
  \item The proposed semantic based feature clustering effectively controls the model complexity. 
  \item Compare to existing feature extraction and topic modeling approach, the proposed model improves the accuracy of the topic prediction by up to 12.99\%.
\end{itemize}

The rest of the paper is structured as follows: In Section \ref{sec:related}, we review the existing methods, from which we got the inspirations. This is followed in Section \ref{sec:approach} by details about our topic models. Section \ref{sec:eval} describes our experimental steps and analyzes the results. Finally, Section \ref{sec:conclusion} concludes this work.

\section{Related Work}\label{sec:related} 

Many topic models have been proposed in the past decades. This includes LDA, Latent Semantic Analysis(LSA), word2vec, and RBM, etc. In this section, we will compare the pros and cons of these topic models for their performance in topic modeling.

LDA was one of the most widely used topic models. LDA introduces sparse Dirichlet prior distributions over document-topic and topic-word distributions, encoding the intuition that documents cover a small number of topics and that topics often use a small number of words \cite{blei2003latent}. LSA was another topic modeling technique which is frequently used in information retrieval. LSA learned latent topics by performing a matrix decomposition (SVD) on the term-document matrix  \cite{dumais2004latent}. In practice, training the LSA model is faster than training the LDA model, but the LDA model is more accurate than the LSA model.

Traditional topic models did not consider the semantic meaning of each word and cannot represent the relationship between different words. Word2vec can be used for learning high-quality word vectors from huge data sets with billions of words, and with millions of words in the vocabulary \cite{mikolov2013efficient}. During the training, the model generated word-context pairs by applying a sliding window to scan through a text corpus. Then the word2vec model trained word embeddings using word-context pairs by using the continuous bag of words (CBOW) model and the skip-gram model \cite{NIPS2013_5021}. The generated word vectors can be summed together to form a semantically meaningful combination of both words. 

RBM was proposed to extract low-dimensional latent semantic representations from a large collection of documents \cite{hinton2009replicated}. The architecture of the RBM is an undirected bipartite graphic, in which word-count vectors are modeled as Softmax input units and the output units are binary units. The Contrastive Divergence learning was used to approximate the gradient. By running the Gibbs sampler, the RBM reconstructed the distribution of units \cite{salakhutdinov2007restricted}. A deeper structure of neural network, the Deep Belief Network (DBN), was developed based on stacked RBMs. 
In \cite{hinton2011discovering}, the input layer was the same as RBM mentioned above, other layers are all binary units.

In this work, we adopt Restricted Boltzmann Machine (RBM) for topic modeling, and investigate feature selection for this model. Another state-of-the-art model in topic modeling is the LDA model. As mentioned in Section \ref{sec:related}, LDA is a statically model that is widely used for topic modeling. However, previous research \cite{srivastava2013modeling} shows that the RBM based topic modeling gives 5.45\%$\sim$19.94\% higher accuracy than the LDA based model. In Section \ref{sec:eval}, we also compare the MAP score of these two when applied to three different datasets. Our results also show that the RBM model has better efficiency and accuracy than the LDA model. Hence, we focus our discussion only for the RBM based topic modeling.

\section{Approach}\label{sec:approach}

\begin{figure}[htb]
\centering
\includegraphics[width=2.8in]{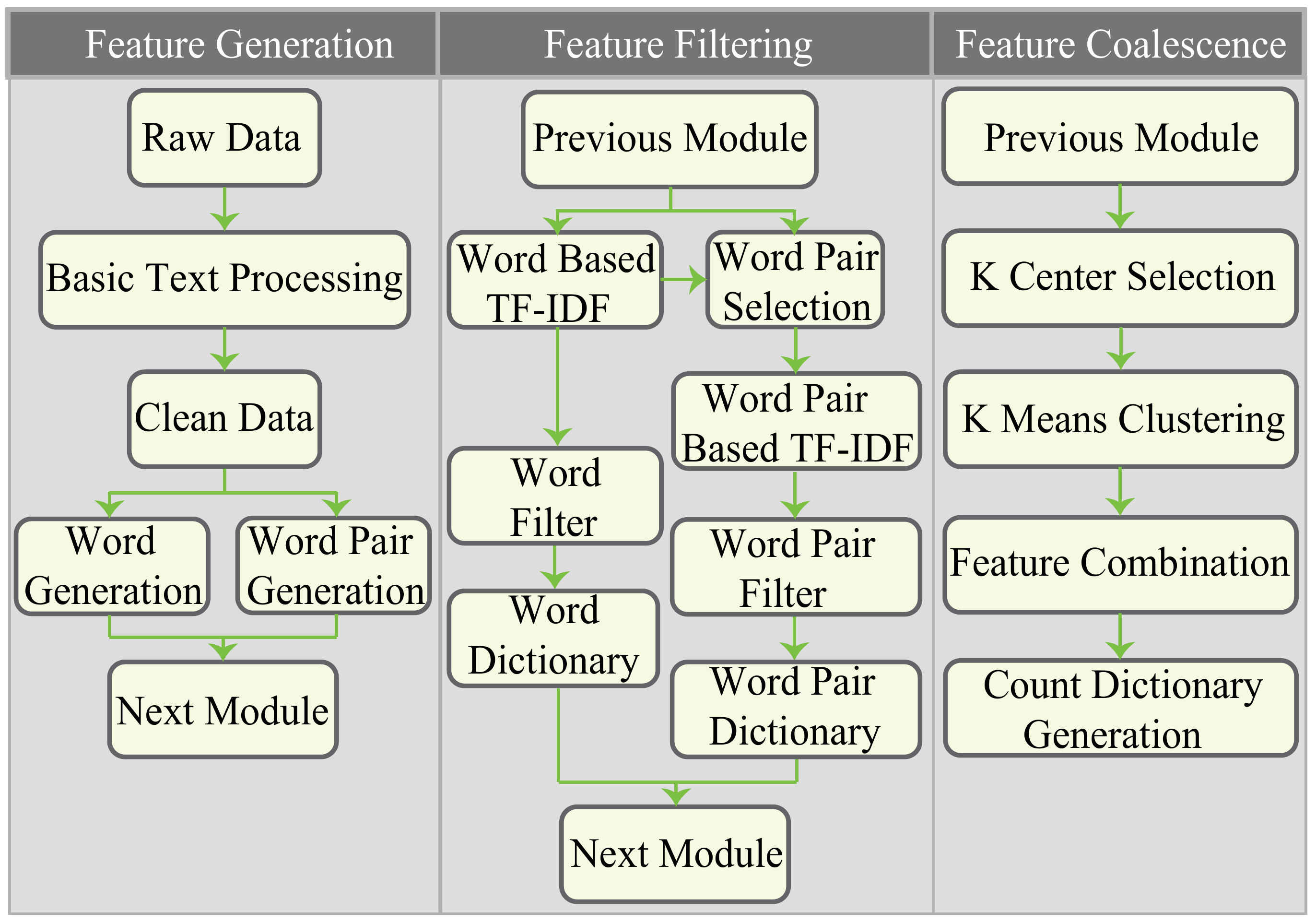}
\caption{Model Structure}
\label{fig:whole_model}
\vspace{-1em}
\end{figure}

Our feature selection contains three steps handled by three different modules: feature generation module, feature filtering module and feature coalescence module. The whole structure of our framework as shown in Figure~\ref{fig:whole_model}. Each module will be elaborated in the next. 

The proposed feature selection is based on our observation that word dependencies provide additional semantic information than that simple word counts. However, there are quadratically more depended word pairs relationships than words. To avoid the explosion of feature set, filtering and coalescing must be performed. Overall those three steps perform feature generation, screening and pooling.

\subsection{Feature Generation: Semantic Word Pair Extraction}

Current RBM model for topic modeling uses the bag-of-words approach. Each visible neuron represents the number of appearance of a dictionary word. We believe that the order of the words also exhibits rich information, which is not captured by the bag-of-words approach. Our hypothesis is that including word pairs (with specific dependencies) helps to improve topic modeling.

\begin{figure}[htb]
\centering
\includegraphics[width=3.4in]{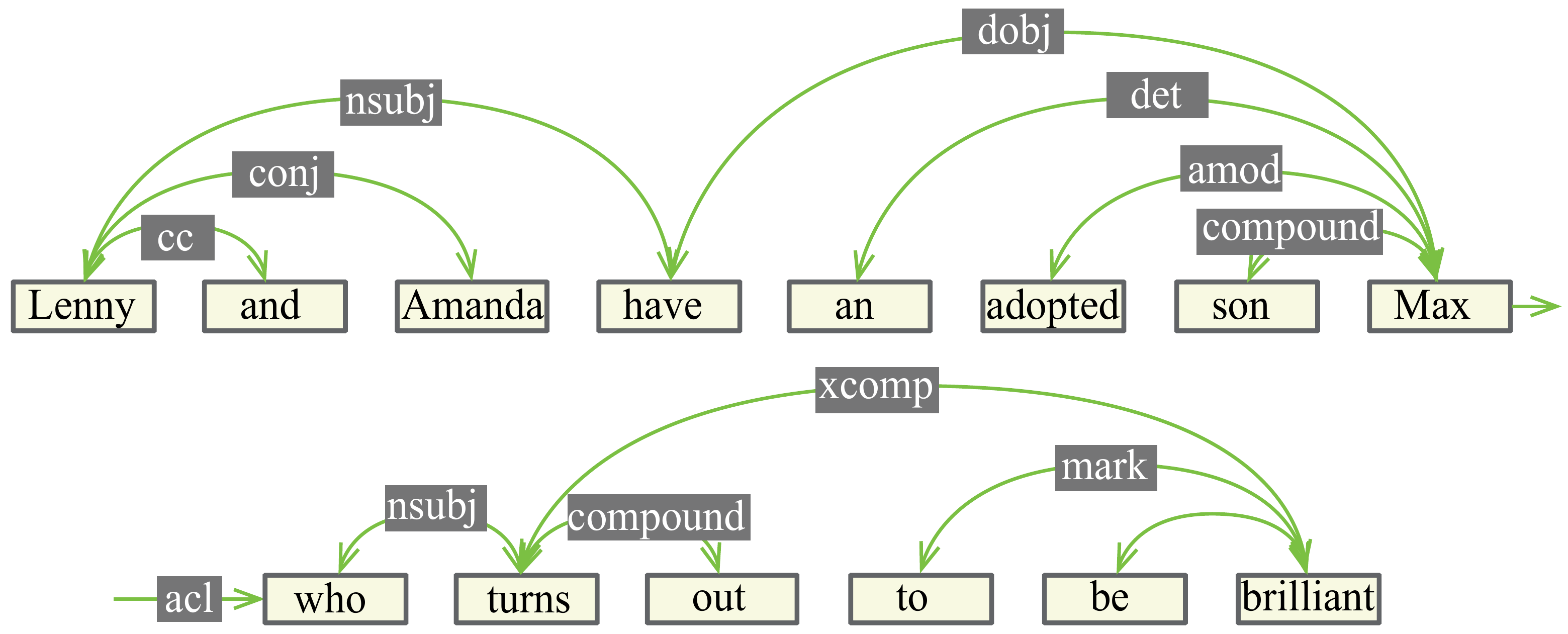}
\caption{Word Pair Extraction}
\label{fig:word_pair}
\vspace{-2em}
\end{figure}

In this work, Stanford natural language parser \cite{chen2014fast} \cite{nivre2016universal} is used to analyze sentences in both training and testing corpus, and extract word pairs that are semantically dependent. Universal dependency(UD) is used during the extraction. For example, given the sentence: \emph{``Lenny and Amanda have an adopted son Max who turns out to be brilliant.''}, which is part of description of the movie ``Mighty Aphrodite'' from the OMDb dataset. Figure~\ref{fig:word_pair} shows all the depended word pairs extracted using the Standford parser. Their order is illustrated by the arrows connection between them, and their relationship is marked beside the arrows. As you can see that the depended words are not necessarily adjacent to each other, however they are semantically related.

Because each single word may have combinations with many other different words during the dependency extraction, the total number of the word pairs will be much larger than the number of word in the training dataset. If we use all depended word pairs extracted from the training corpus, it will significantly increase the size of our dictionary and reduce the performance. To retain enough information with manageable complexity, we keep the 10,000 most frequent word pairs as the initial word pair dictionary. Input features of the topic model will be selected from this dictionary. Similarly, we use the 10,000 most frequent words to form a word dictionary. For both dictionary, stop words are removed.

\subsection{Feature Filtering: Two steps TF-IDF Processing}

The word dictionary and word pair dictionary still contain a lot of high frequency words that are not very informational, such as "first", "name", etc. \emph{Term frequency-inverse document frequency (TF-IDF)} is applied to screen out those unimportant words or word pairs and keep only important ones. The equation to calculate TF-IDF weight is as following:

\begin{equation} \label{eq:term_frequency}
\scalebox{1.0}{$TF(t) = \frac{\mathrm{Number\ of\ times\ term\ \textit{t}\ appears\ in\ a\ document}}{\mathrm{Total\ number\ of\ terms\ in\ the\ document}}$}
\end{equation}

\begin{equation} \label{eq:inverse_document_frequency}
\scalebox{1.0}{$IDF(t) = \log\frac{\mathrm{Total\ number\ of\ documents}}{\mathrm{Number\ of\ documents\ with\ term\ \textit{t}\ in\ it}}$}
\end{equation}

\begin{equation} \label{eq:tfidf}
\scalebox{1.0}{$TF-IDF(t) = TF(t)*IDF(t)$}
\end{equation}

Equation~\ref{eq:term_frequency} calculates the \emph{Term Frequency (TF)}, which measures how frequently a term occurs in a document. Equation~\ref{eq:inverse_document_frequency} calculates the \emph{Inverse document frequency (IDF)}, which measures how important a term is. The TF-IDF weight is often used in information retrieval and text mining. It is a statically measure to evaluate how important a word is to a document in a collection of corpus. The importance increases proportionally to the number of times a word appears in the document but is offset by the frequency of the word in the corpus \cite{sparck1972statistical} \cite{salton1983extended} \cite{salton1986introduction} \cite{salton1988term} \cite{wu2008interpreting}. 

As shown in Figure~\ref{fig:whole_model}, Feature Filtering module, a two-step TF-IDF processing is adopted. First, the word-level TF-IDF is performed. The result of word level TF-IDF is used as a filter and a word pair is kept only if the TF-IDF scores of both words are higher than the threshold (0.01).  After that, we treat each word pair as a single unit, and the TF-IDF algorithm is applied again to the word pairs and further filter out word pairs that are either too common or too rare. Finally, this module will generate the filtered word dictionary and the filtered word pair dictionary.

\subsection{Feature Coalescence: K-means Clustering}

Even with the TF-IDF processing, the size of the word pair dictionary is still prohibitively large. We further cluster semantically close word pairs to reduce the dictionary size. Each word is represented by their embedded vectors calculated using Google's word2vec model. The semantic distance between two words is measured as the Euclidean distance of their embedding vectors. The words that are semantically close to each other are grouped into K clusters.

We use the index of each cluster to replace the words in the word pair. If the cluster ID of two word pairs are the same, then the two word pairs are semantically similar and be merged. In this we can reduce the number of word pairs by more than 63\%. We also investigate how the number of the cluster centrum (i.e. the variable K) will affect the model accuracy. The detailed experimental results on three different datasets will be given in \ref{sec:eval}.

\section{Evaluation}\label{sec:eval}
\subsection{Experiments Setup} 
The proposed topic model will be tested in the context of content-based recommendation. Given a query document, the goal is to search the database and find other documents that fall into the category by analyzing their contents. In our experiment, we generate the topic distribution of each document by using RBM model. Then we retrieve the top N documents whose topic is the closet to the query document by calculating their Euclidean distance. The number of hidden units of the RBM is 500 which represents 500 topics. The number of visible units of the RBM equals to total number of different words and words pairs extracted as input features. The weights are updated using a learning rate of 0.01. During the training, momentum, epoch, and weight decay are set to be 0.9, 15, and 0.0002 respectively.

Our proposed method is evaluated on 3 datasets: OMDb, Reuters, and 20NewsGroup. All the datasets are divided into three subsets: training, validation, and testing. The split ratio is 70:10:20. For each dataset, a 5-fold cross-validation is applied.

\begin{itemize}
\item OMDb, the Open Movie Database, is a database of movie information. The OMDb dataset is collected using OMDb APIs \cite{fritzomdb}. The training dataset contains 6043 movie descriptions; the validation dataset contains 863 movie descriptions and the testing dataset contains 1727 movie descriptions. Based on the genre of the movie, we divided them into 20 categories and tagged them accordingly.

\item The Reuters, is a dataset consists of documents appeared on the Reuters newswire in 1987 and were manually classified into 8 categories by personnel from Reuters Ltd. There are 7674 documents in total. The training dataset contains 5485 news, the validation dataset contains 768 news and the testing dataset contains 1535 news.

\item The 20NewsGroup dataset is a collection of approximately 20,000 newsgroup documents, partitioned (nearly) evenly across 20 different newsgroups. The training dataset contains 13174 news, the validation dataset contains 1882 news and the testing dataset contains 3765 news. Both Reuters and 20NewsGroup dataset are download from \cite{cachopo2007improving}.
\end{itemize}

\subsection{Metric}

We use \begin{math} mean\ Average\ Precision\ (mAP) \end{math} score to evaluate our proposed method. It is a score to evaluate the information retrieval quality. This evaluation method considers the effect of orders in the information retrieval results. A higher \begin{math} mAP \end{math} score is better. If the relational result is shown in the front position (i.e. ranks higher in the recommendation), the score will be close to 1; if the relational result is shown in the back position (i.e. ranks lower in the recommendation), the score will be close to 0. \begin{math} mAP1 \end{math}, \begin{math} mAP3 \end{math}, \begin{math} mAP5 \end{math}, and \begin{math} mAP10 \end{math} are used to evaluate the retrieval performance. For each document, we retrieve 1, 3, 5, and 10 documents whose topic vectors have the smallest Euclidean distance with that of the query document. The documents are considered as relevant if they share the same class label. Before we calculate the \begin{math} mAP \end{math}, we need to calculate the \begin{math} Average\ Precision\ (AveP)  \end{math} for each document first. The equation of \begin{math} AveP  \end{math} is described below,
\begin{equation} \label{eq:average_precision}
AveP = \frac{\sum_{k=1}^{n}(P(k)\cdot rel(k))}{\textrm{number of relevant documents}},
\end{equation}
where \begin{math} rel(k) \end{math} is an indicator function equaling $1$ if the item at rank \begin{math} k \end{math} is a relevant document, $0$ otherwise \cite{turpin2006user}. Note that the average is over all relevant documents and the relevant documents not retrieved get a precision score of zero.

The equation of the \begin{math} mean\ Average\ Precision\ (mAP)  \end{math} score is as following,
\begin{equation} \label{eq:map_score}
mAP = \frac{\sum_{q=1}^{Q}AveP(q)}{Q},
\end{equation}
where \begin{math} Q \end{math} indicates the total number of queries.

\subsection{Results}

\subsubsection{LDA and RBM Performance Comparison}

In the first experiment, we investigate the topic modeling performance between LDA and RBM. For the training of the LDA model, the training iteration is 15 and the number of generated topics is 500 which are as the same as the RBM model. As we can see from the Table~\ref{LDA/RBM-table}. The RBM outperforms the LDA in all datasets. For example, using the \begin{math} mAP5 \end{math} evaluation, the RBM is 30.22\% greater than the LDA in OMDb dataset, 18.18\% greater in Reuters dataset and 25.25\% greater in 20NewsGroup dataset. To have a fair comparison, the RBM model here is based on word only features. In the next we will show the including word pairs can further improve its \begin{math} mAP \end{math} score.

\begin{table}[htb]
\caption{LDA and RBM performance evaluation}
\label{LDA/RBM-table}
\centering
\resizebox{1.0\columnwidth}{!}{
\begin{tabular}{|c|*{7}{c|}}
\hline
\multirow{2}{*}{\bf mAP }  &
\multicolumn{2}{c|}{\bf OMDb} &
\multicolumn{2}{c|}{\bf Reuters} &
\multicolumn{2}{c|}{\bf 20NewsGroup} \\
& LDA & RBM & LDA & RBM & LDA & RBM  \\ 
\hline
mAP 1      &0.12166      &\textbf{0.14772}     &0.84919    &\textbf{0.94407}    &0.68669    &\textbf{0.73959}\\
\hline 
mAP 3      &0.07473      &\textbf{0.09381}     &0.79976    &\textbf{0.92604}    &0.55410    &\textbf{0.65530}\\
\hline 
mAP 5      &0.05723      &\textbf{0.07453}     &0.77500    &\textbf{0.91589}    &0.48796    &\textbf{0.61115}\\
\hline 
mAP 10     &0.03914      &\textbf{0.05273}     &0.74315    &\textbf{0.90050}    &0.44719    &\textbf{0.55338}\\
\hline
\end{tabular}
}
\vspace{-2em}
\end{table}

\subsubsection{Word/Word Pair Performance Comparison}

In this experiment, we compare the performance of two RBM models. One of them only considers words as the input feature, while the other has combined words and word pairs as the input feature. The  total feature size varies from 10500, 11000, 11500, 12000, 12500, 15000. For the word/word pair combined RBM model, the number of word feature is fixed to be 10000, and the number of word pair features is set to meet the requirement of total feature size. 


\begin{table*}[!t]
\caption{fixed total feature number word/word pair performance evaluation}
\label{Word/Word_Pair-Feature-table}
\centering
\resizebox{2.0\columnwidth}{!}{
\begin{tabular}{|c|*{13}{c|}}
\hline
\multirow{2}{*}{\bf mAP }  &
\multicolumn{2}{c|}{\bf F = 10.5K} &
\multicolumn{2}{c|}{\bf F = 11K} &
\multicolumn{2}{c|}{\bf F = 11.5K} & 
\multicolumn{2}{c|}{\bf F = 12K} &
\multicolumn{2}{c|}{\bf F = 12.5K} &
\multicolumn{2}{c|}{\bf F = 15K} \\
& word & word pair & word & word pair & word & word pair & word & word pair & word & word pair & word & word pair
\\ \hline 
\multicolumn{13}{|c|}{\bf OMDB }\\
\hline
mAP 1      &\textbf{0.14772}      &0.14603     &0.13281    &\textbf{0.14673}    &0.13817    &\textbf{0.14789}    &0.13860      &\textbf{0.14754}     &0.14019    &\textbf{0.14870}    &0.13686    &\textbf{0.14708}\\
\hline 
mAP 3      &0.09381      &\textbf{0.09465}     &0.08606    &\textbf{0.09327}    &0.08933    &\textbf{0.09507}    &0.08703      &\textbf{0.09517}     &0.09054    &\textbf{0.09657}    &0.09009    &\textbf{0.09537}\\
\hline 
mAP 5      &0.07453      &\textbf{0.07457}     &0.06835    &\textbf{0.07380}    &0.07089    &\textbf{0.07508}    &0.06925      &\textbf{0.07485}     &0.07117    &\textbf{0.07635}    &0.07175    &\textbf{0.07511}\\
\hline 
mAP 10     &0.05273      &\textbf{0.05387}     &0.04862    &\textbf{0.05340}    &0.04976    &\textbf{0.05389}    &0.04900      &\textbf{0.05322}     &0.05019    &\textbf{0.05501}    &0.05083    &\textbf{0.05388}\\
\hline
\multicolumn{13}{|c|}{\bf Reuters }\\
\hline
mAP 1      &0.94195      &\textbf{0.95127}     &0.94277    &\textbf{0.95023}    &0.94407    &\textbf{0.95179}    &0.94244      &\textbf{0.94997}     &0.94277    &\textbf{0.95270}    &0.94163    &\textbf{0.94984}\\
\hline 
mAP 3      &0.92399      &\textbf{0.93113}     &0.92448    &\textbf{0.93117}    &0.92604    &\textbf{0.93276}    &0.92403      &\textbf{0.93144}     &0.92249    &\textbf{0.93251}    &0.92326    &\textbf{0.93353}\\
\hline 
mAP 5      &0.91367      &\textbf{0.92123}     &0.91366    &\textbf{0.91939}    &0.91589    &\textbf{0.92221}    &0.91367      &\textbf{0.92051}     &0.91310    &\textbf{0.92063}    &0.91284    &\textbf{0.92219}\\
\hline 
mAP 10     &0.89813      &\textbf{0.90425}     &0.89849    &\textbf{0.90296}    &0.90050    &\textbf{0.90534}    &0.89832      &\textbf{0.90556}     &0.89770    &\textbf{0.90365}    &0.89698    &\textbf{0.90499}\\
\hline 
\multicolumn{13}{|c|}{\bf 20NewsGroup }\\
\hline
mAP 1      &0.73736      &\textbf{0.77129}     &0.73375    &\textbf{0.76093}    &0.68720    &\textbf{0.75865}    &0.73959      &\textbf{0.75846}     &0.72280    &\textbf{0.76768}    &0.72695    &\textbf{0.75583}\\
\hline 
mAP 3      &0.65227      &\textbf{0.68905}     &0.64848    &\textbf{0.68042}    &0.60356    &\textbf{0.67546}    &0.65530      &\textbf{0.67320}     &0.63649    &\textbf{0.68455}    &0.63951    &\textbf{0.66743}\\
\hline 
mAP 5      &0.60861      &\textbf{0.64620}     &0.60548    &\textbf{0.63783}    &0.56304    &\textbf{0.63321}    &0.61115      &\textbf{0.62964}     &0.59267    &\textbf{0.64165}    &0.59447    &\textbf{0.62593}\\
\hline 
mAP 10     &0.55103      &\textbf{0.58992}     &0.54812    &\textbf{0.58057}    &0.51188    &\textbf{0.57839}    &0.55338      &\textbf{0.57157}     &0.53486    &\textbf{0.58500}    &0.53749    &\textbf{0.56969}\\
\hline
\end{tabular}
}
\vspace{-3em}
\end{table*}

Both models are first applied to the OMDb dataset, and the results are shown in Table~\ref{Word/Word_Pair-Feature-table}, section 1, the word/word pair combined model almost always performs better than the word-only model. For the \begin{math} mAP1 \end{math}, the \begin{math} mAP5 \end{math} and the \begin{math} mAP10 \end{math}, the most significant improvement occurs when total feature size is set to = 11000. About 10.48\%, 7.97\%, and 9.83\% improved were found compared to the word-only model. For the \begin{math} mAP3 \end{math}, the most significant improvement occurs when the total feature size is set to = 12000, and about 9.35\% improvement is achieved by considering word pair.

The two models are further applied on the Reuters dataset, and the results are shown in Table~\ref{Word/Word_Pair-Feature-table}, section 2. Again, the word/word pair combined model outperforms the word-only model almost all the time. For the \begin{math} mAP1, 3, 5\ and\ 10 \end{math} up to 1.05\%, 1.11\%, 1.02\% and 0.89\% improvement are achieved.

The results for 20NewsGroup dataset are shown in  Table~\ref{Word/Word_Pair-Feature-table}, section 3. Similar to previous two datasets, all the results from word/word pair combined model are better than the word-only model. For the \begin{math} mAP1, 3, 5\ and\ 10 \end{math}, the most significant improvement occurs when when the total feature size is set to = 11500. Up to 10.40\%, 11.91\%, 12.46\% and 12.99\% improvements can achieved.

\subsubsection{Cluster Centrum Selection}

In the third experiment, we focus on how the different K values affect the effectiveness of the generated word pairs in terms of their ability of topic modeling. The potential K values are 100, 300, 500, 800 and 1000. Then we compare the \begin{math} mAP \end{math} between our model and the baseline model, which consists of word only input features.

\begin{figure}[htb] 
  \centering 
  \subfloat[OMDb mAP1]{\includegraphics[width=1.7in]{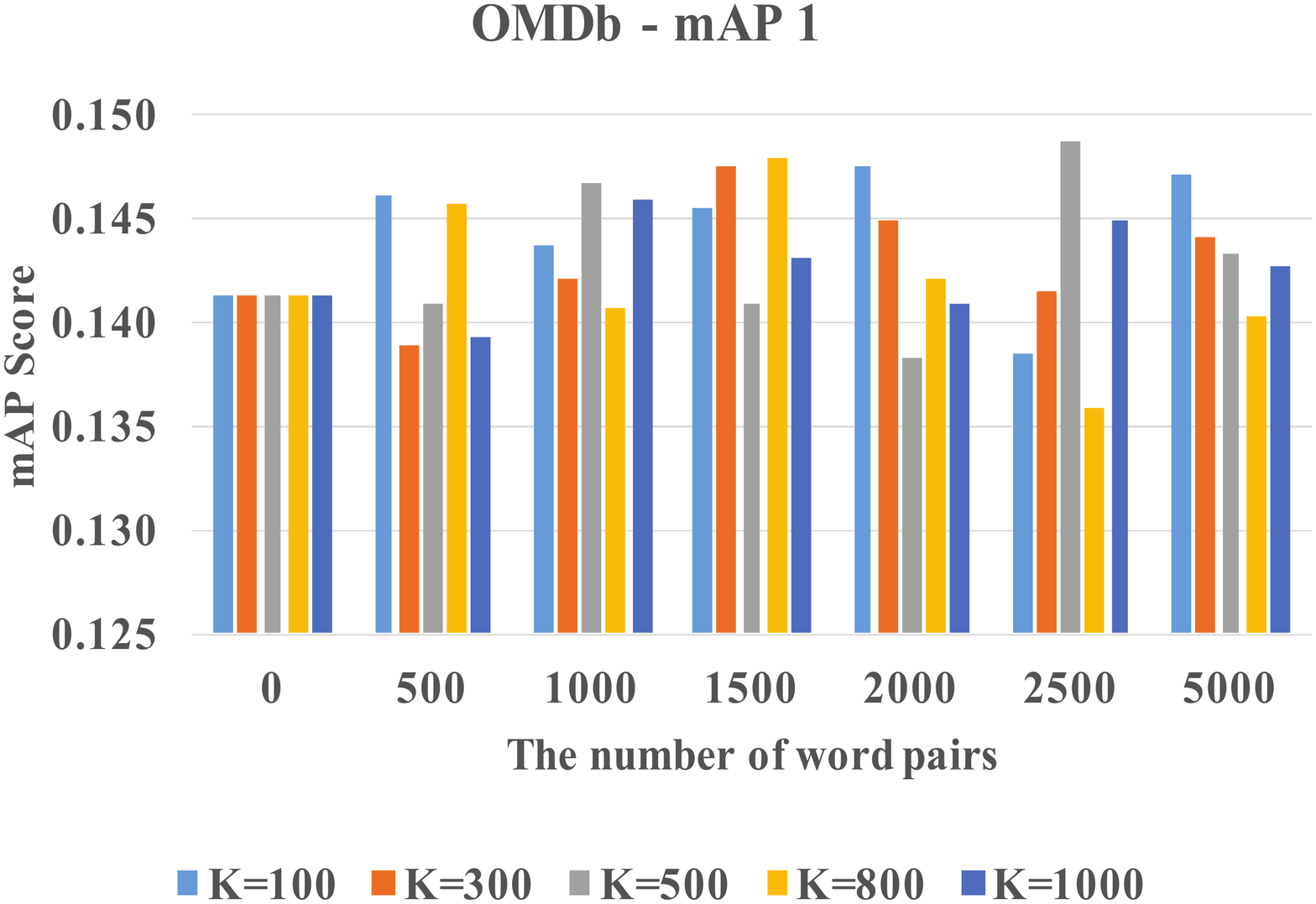}%
  \label{OMDb-mAP1}} 
  \hfil
  \subfloat[OMDb mAP3]{\includegraphics[width=1.7in]{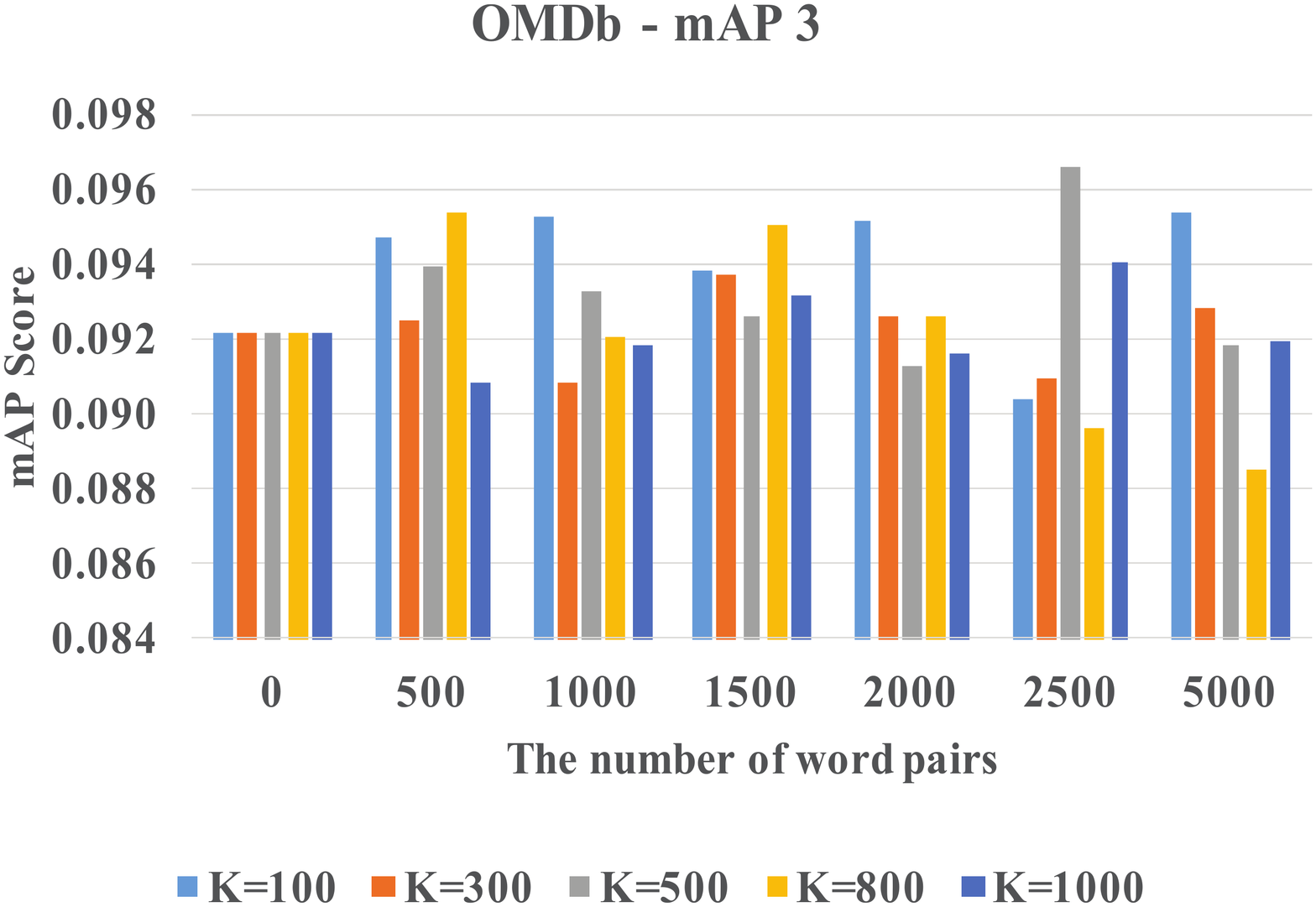}%
  \label{OMDb-mAP3}} 
  \hfil
  \subfloat[OMDb mAP5]{\includegraphics[width=1.7in]{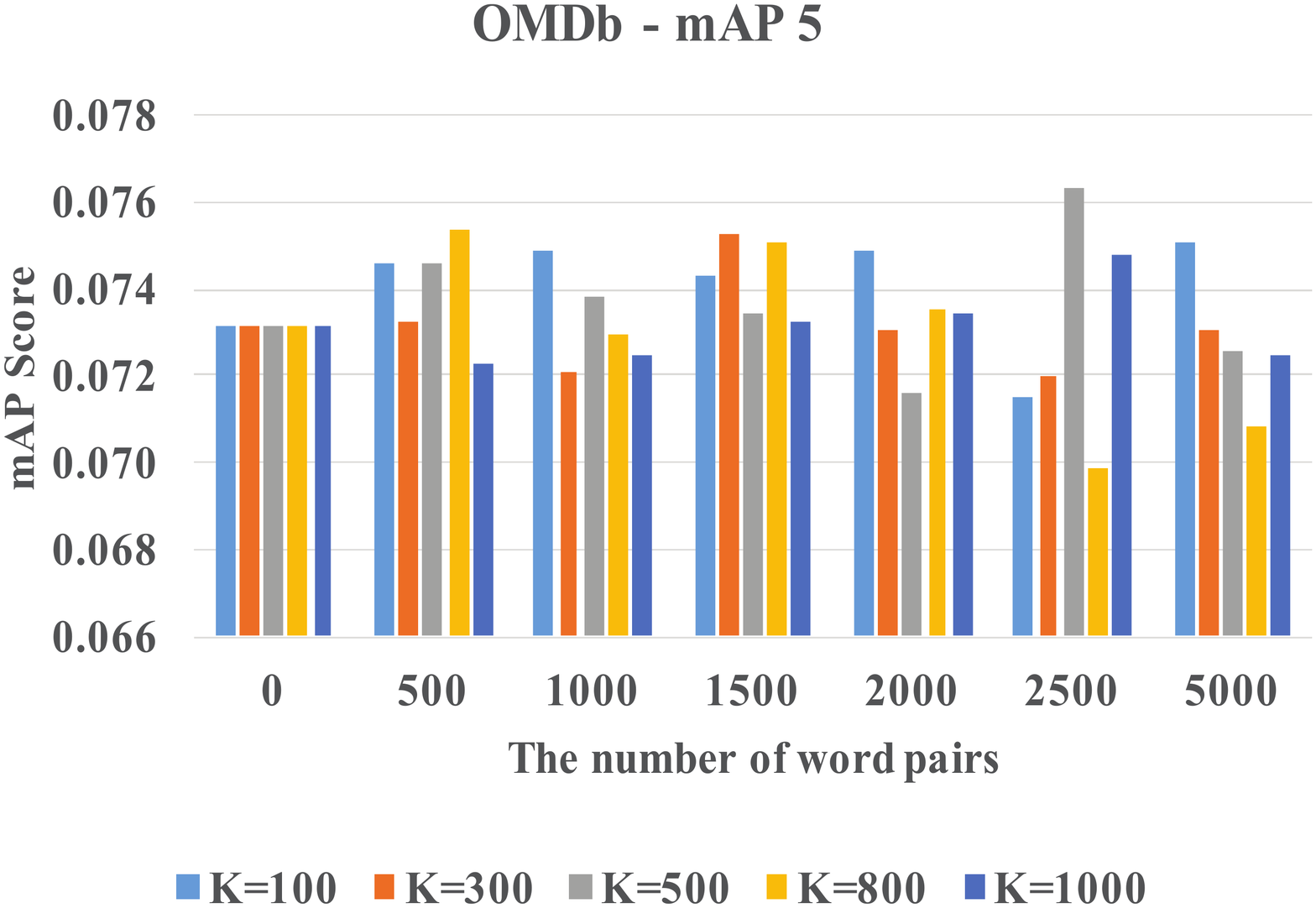}%
  \label{OMDb-mAP5}} 
  \hfil
  \subfloat[OMDb mAP10]{\includegraphics[width=1.7in]{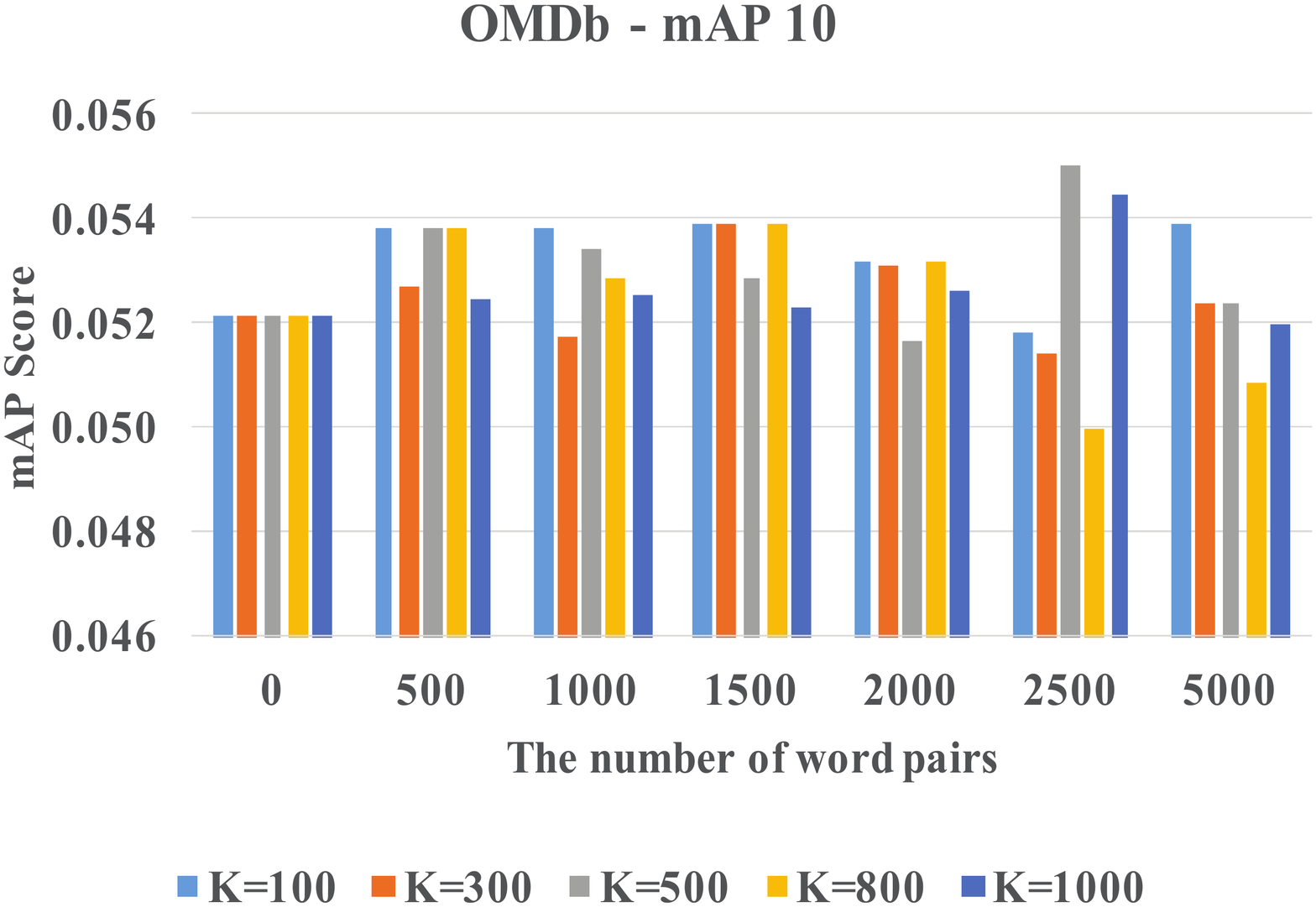}%
  \label{OMDb-mAP10}} 
  \caption{OMDb dataset mAP score evaluation} 
  \label{OMDb-mAP} 
  \vspace{-2em}
\end{figure}

The OMDb dataset results are shown in Figure~\ref{OMDb-mAP}. As we can observe, all the K values give us better performance than the baseline. The most significant improvement occurs when in K = 100. Regardless of the size of word pair features, in average we can achieve 2.41\%, 2.15\%, 1.46\% and 4.46\% improvements in \begin{math} mAP1, 3, 5, and\ 10\end{math} respectively.

\begin{figure}[htb] 
  \centering 
  \subfloat[Reuters mAP1]{\includegraphics[width=1.7in]{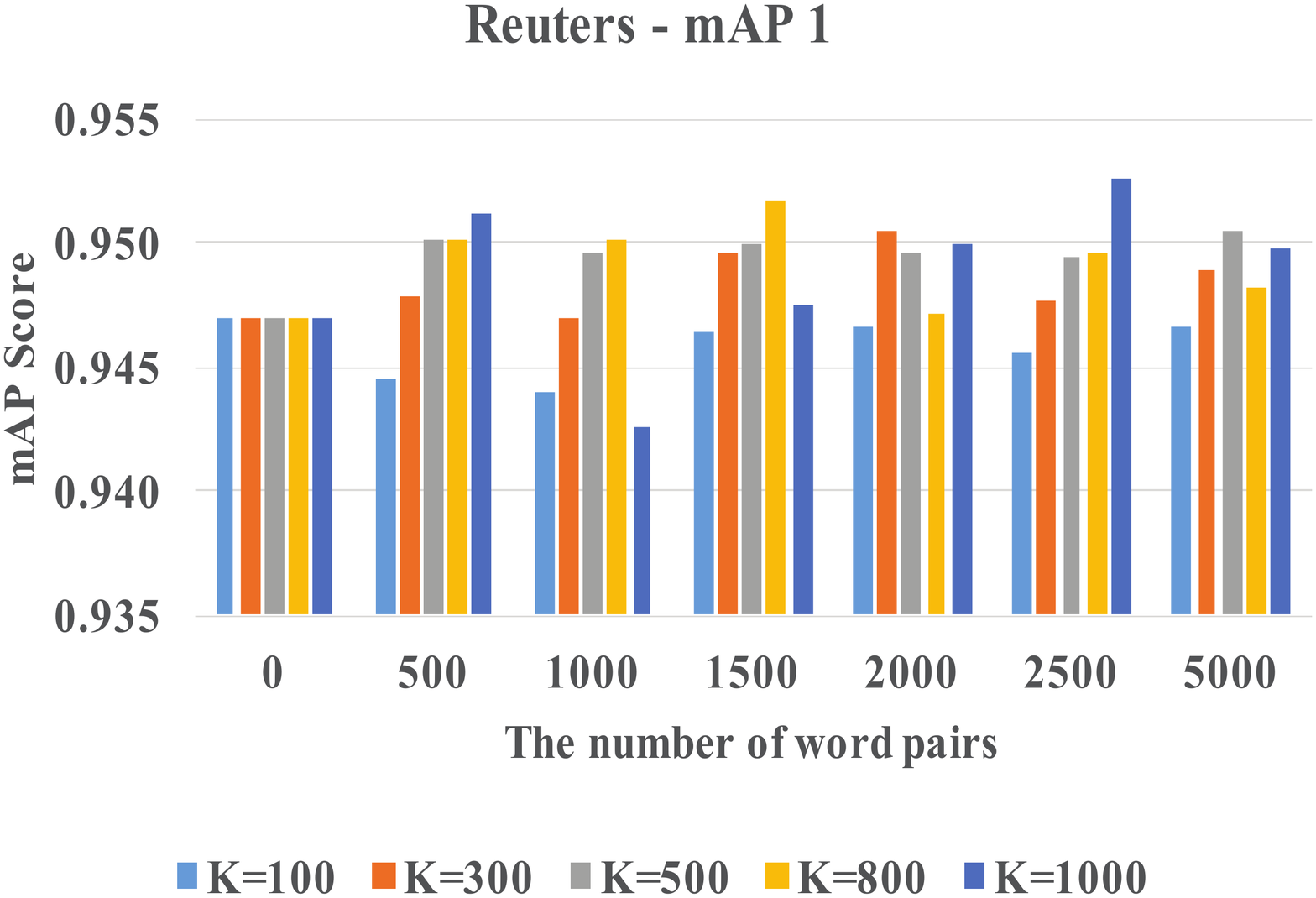}%
  \label{Reuters-mAP1}} 
  \hfil
  \subfloat[Reuters mAP3]{\includegraphics[width=1.7in]{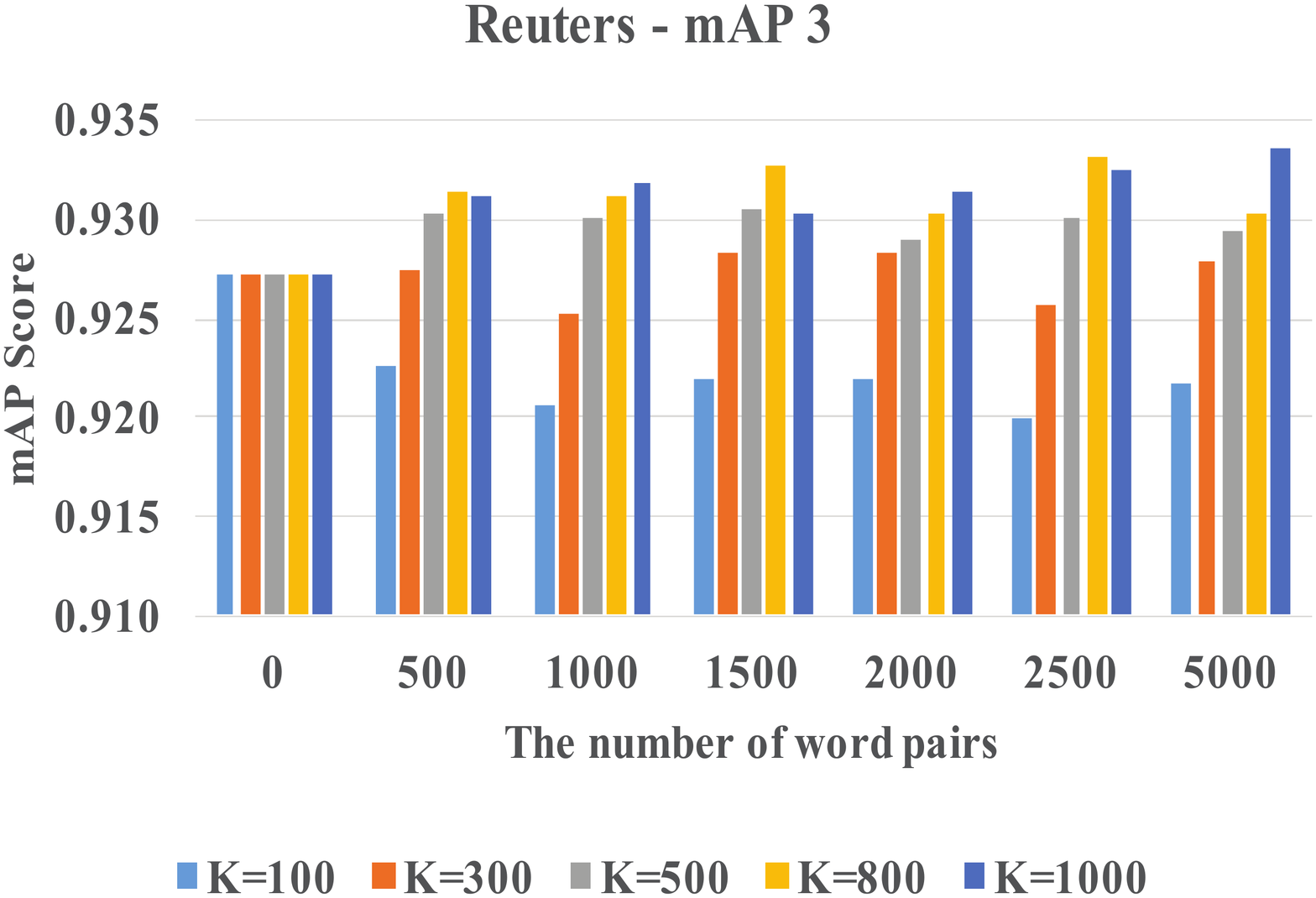}%
  \label{Reuters-mAP3}} 
  \hfil
  \subfloat[Reuters mAP5]{\includegraphics[width=1.7in]{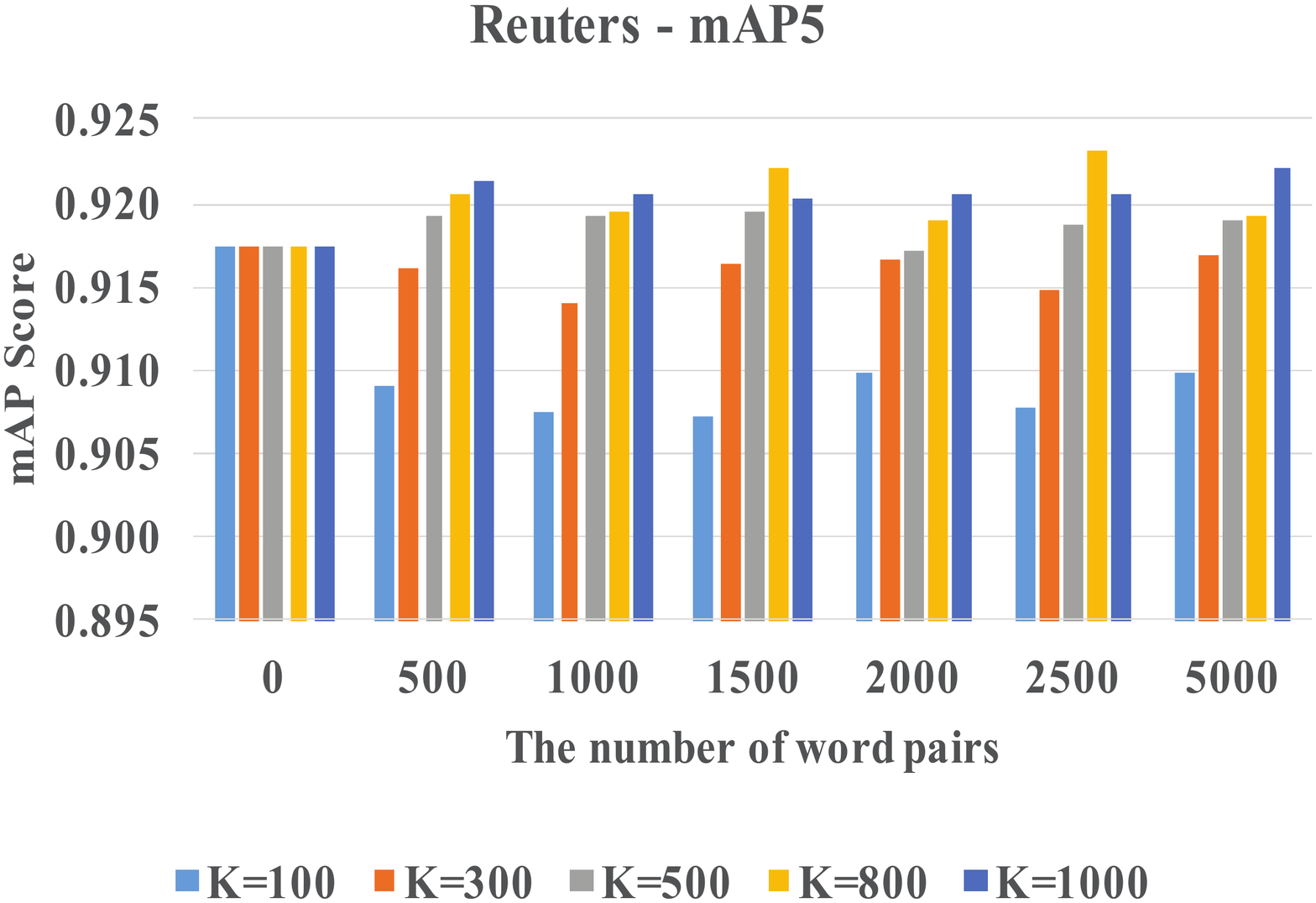}%
  \label{Reuters-mAP5}} 
  \hfil
  \subfloat[Reuters mAP10]{\includegraphics[width=1.7in]{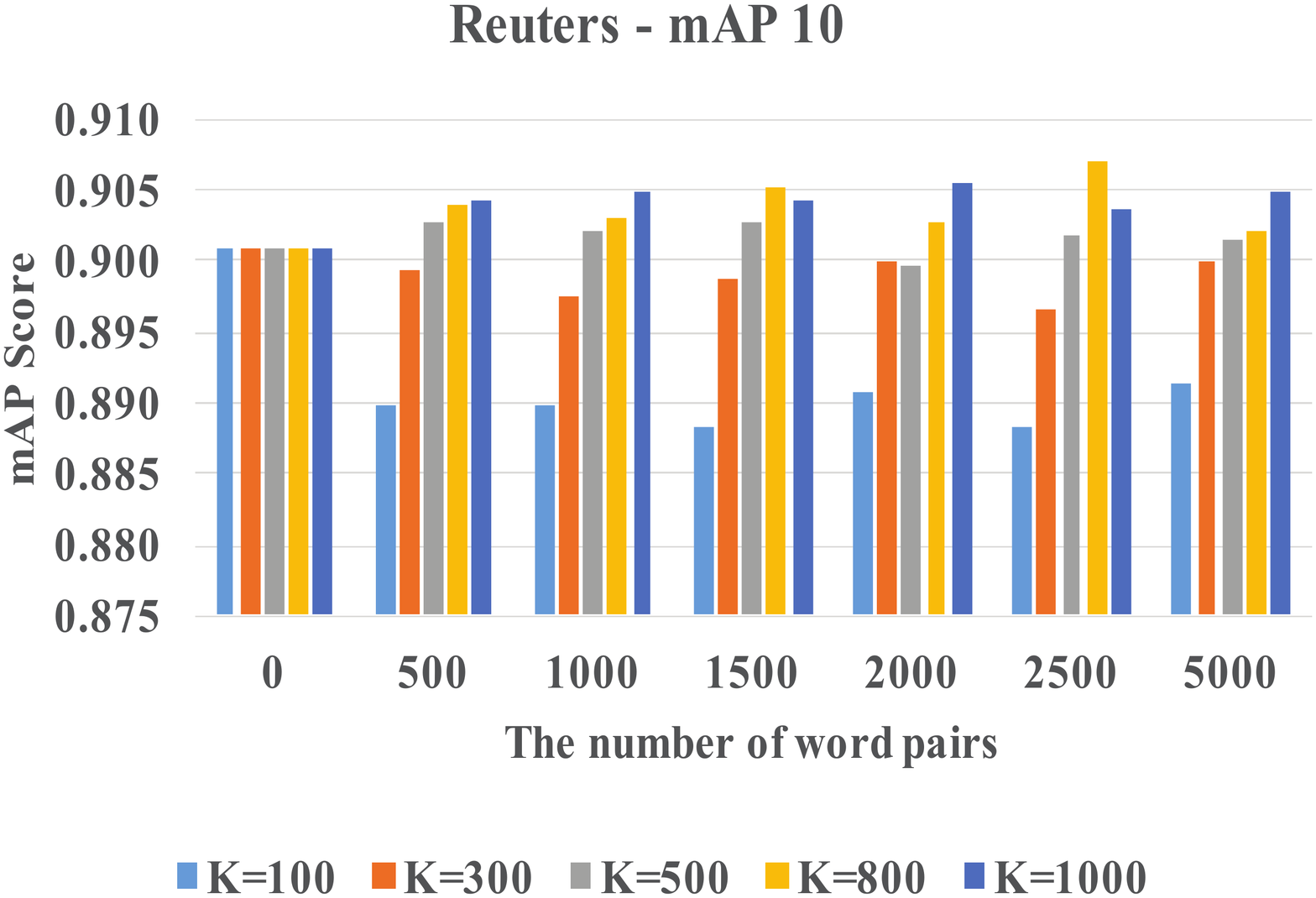}%
  \label{Reuters-mAP10}} 
  \caption{Reuters dataset mAP score evaluation} 
  \label{Reuters-mAP} 
  \vspace{-2em}
\end{figure}

The results of Reuters dataset are shown in Figure~\ref{Reuters-mAP}. When the K value is greater than 500, all \begin{math} mAP \end{math} scores for word/word pair combination model are better than the baseline. Because the mAP score for Reuters dataset in original model is already very high (almost all of them are higher than 0.9), compared to OMDb, it is more difficult to further improve the \begin{math} mAP \end{math} score of this dataset. For the \begin{math} mAP1 \end{math}, disregard the impact of input feature size, in average the most significant improvement happens when K = 500, which is 0.31\%. For the \begin{math} mAP3 \end{math}, the \begin{math} mAP5 \end{math} and the \begin{math} mAP10 \end{math}, the most significant improvements happen when K = 800, which are 0.50\%, 0.38\% and 0.42\% respectively. 

\begin{figure}[htb] 
  \centering 
  \subfloat[20NewsGroup mAP1]{\includegraphics[width=1.7in]{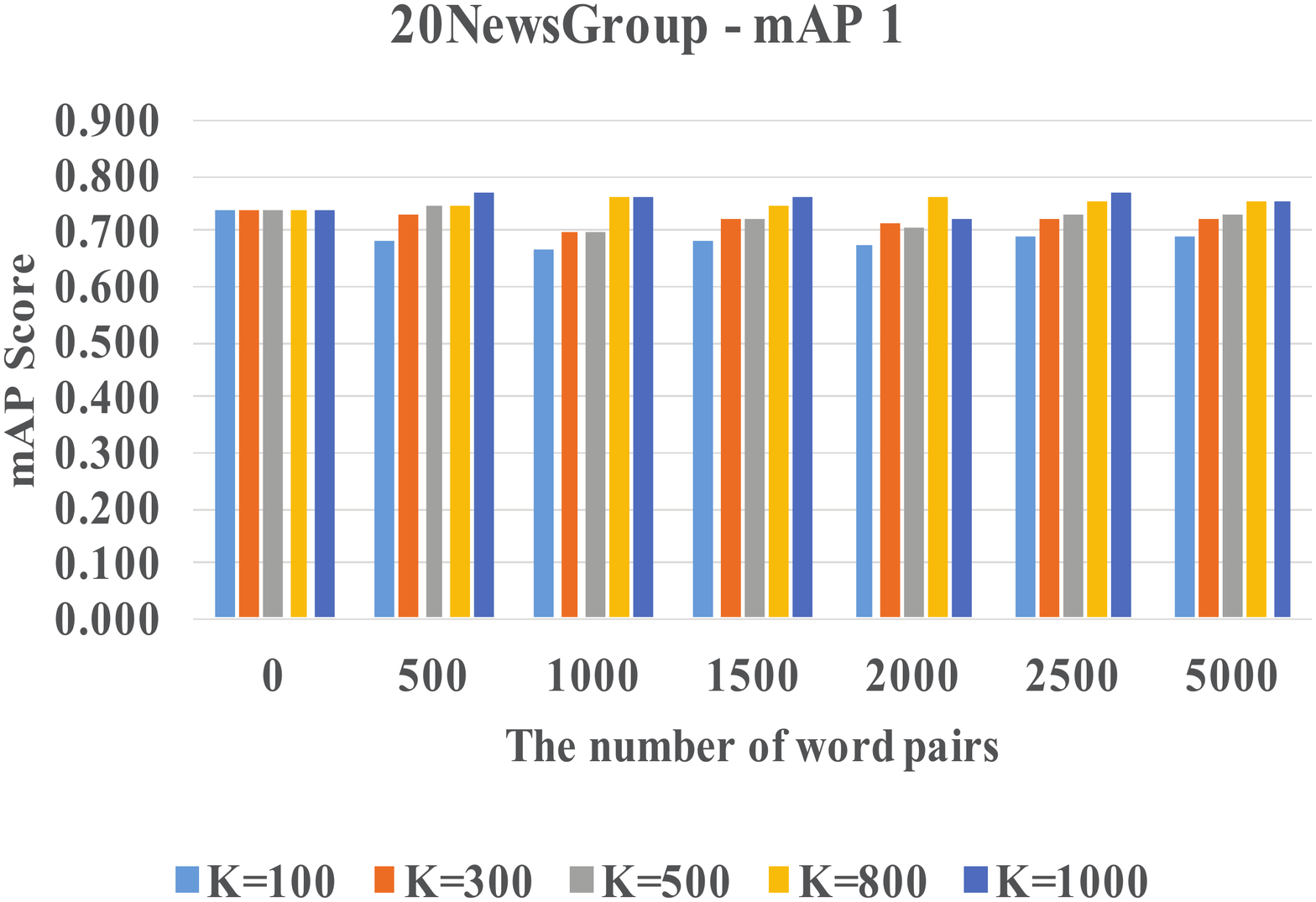}%
  \label{20NewsGroup-mAP1}} 
  \hfil
  \subfloat[20NewsGroup mAP3]{\includegraphics[width=1.7in]{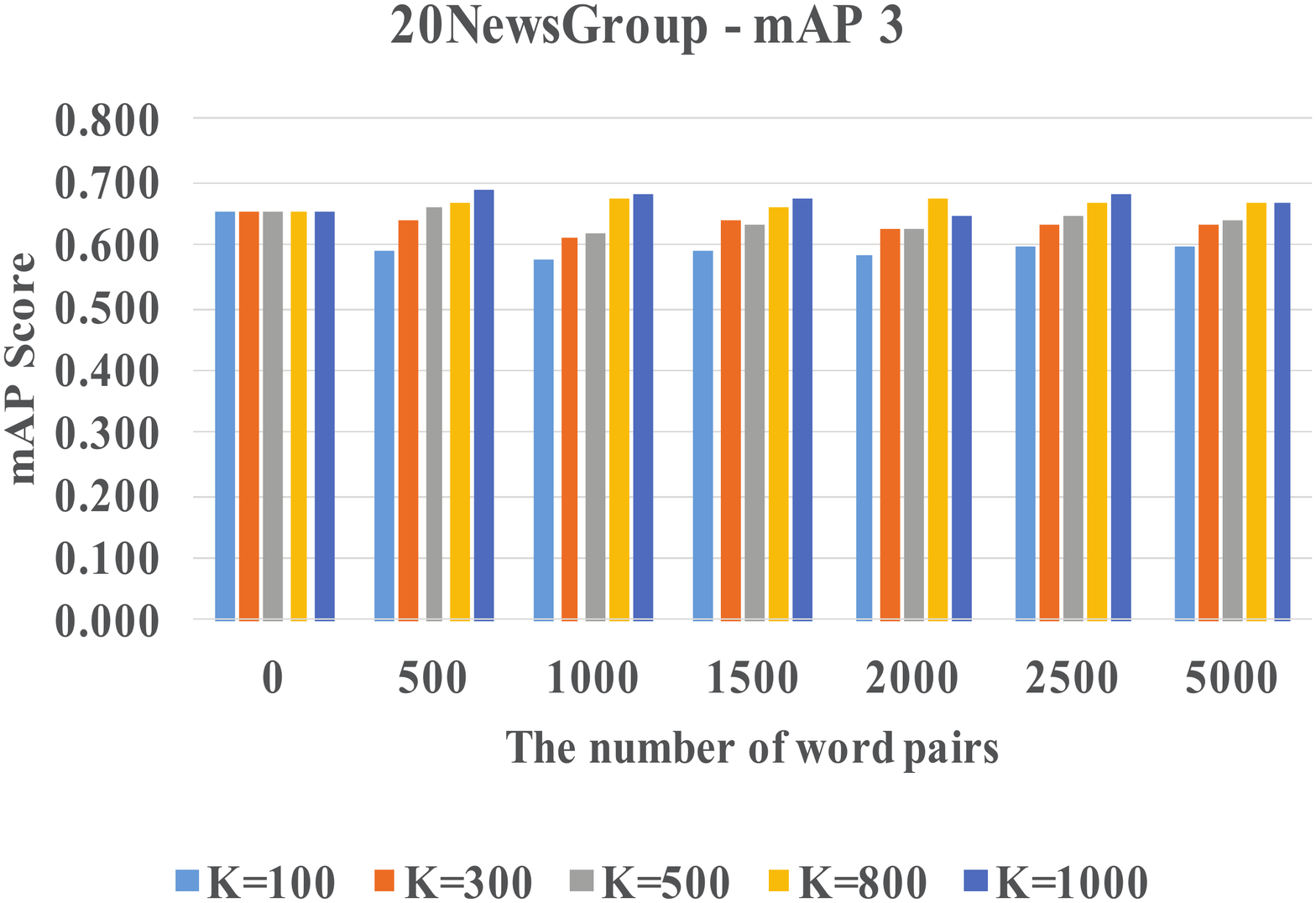}%
  \label{20NewsGroup-mAP3}} 
  \hfil
  \subfloat[20NewsGroup mAP5]{\includegraphics[width=1.7in]{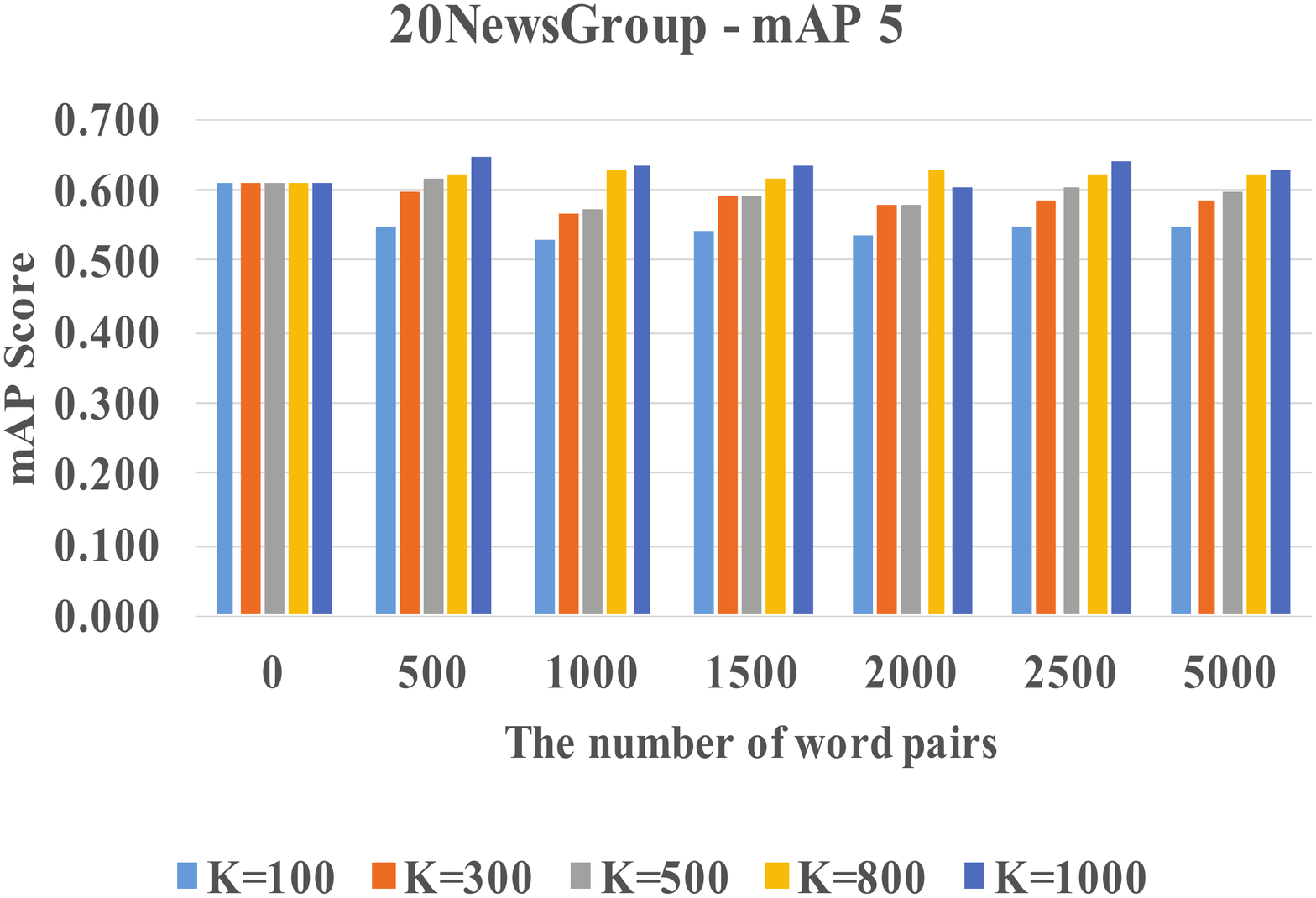}%
  \label{20NewsGroup-mAP5}} 
  \hfil
  \subfloat[20NewsGroup mAP10]{\includegraphics[width=1.7in]{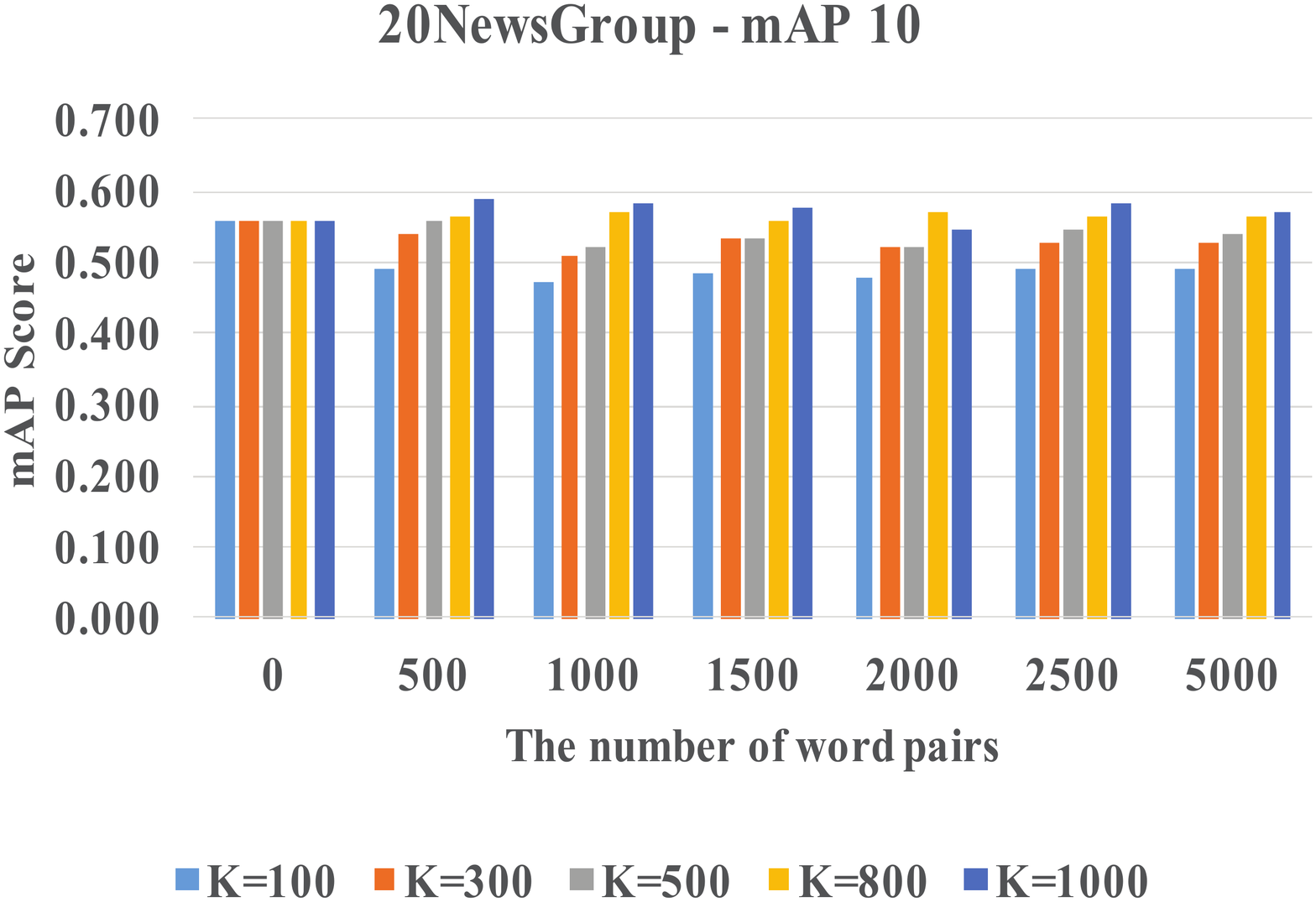}%
  \label{20NewsGroup-mAP10}} 
  \caption{20NewsGroup dataset mAP score evaluation} 
  \label{20NewsGroup-mAP} 
  \vspace{-2em}
\end{figure}

The results for 20NewsGroup dataset results are shown in Figure~\ref{20NewsGroup-mAP}. Similar to the Reuters dataset, when the K value is greater than 800, all \begin{math} mAP \end{math} scores for word/word pair combination model are better than the baseline. For the \begin{math} mAP1, 3, 5, and\ 10\end{math}, in average the most significant improvements are 2.82\%, 2.90\%, 3.2\% and 3.33\% respectively, and they all happen when K = 1000.

In summary, a larger K value generally give a better result, like the Reuters dataset and the 20NewsGroup dataset. However, for some documents sets, such as OMDb, where the vocabulary semantically has a wide distribution, keeping the number of clusters small will not lose too much information. 

\subsubsection{Word Pair Generation Performance}

In the last experiment, we compare different word pair generation algorithms with the baseline. Similar to previous experiments, the baseline is the word-only RBM model whose input consists of the 10000 most frequent words. The ``semantic'' word pair generation is the method we proposed in this paper. The proposed technique is compared to a reference approach that applies the idea from the skip-gram \cite{NIPS2013_5021} algorithm, and generates the word pairs from each word's adjacent neighbor. We call it ``N-gram'' word pair generation. And the window size used in here is N = 2. For the ``Non-K'' word pair generation, we use the same algorithm as the ``semantic'' except that no K-means clustering is applied on the generated word pairs.

\begin{table}[htb]
\caption{Different Word Pair Generation Algorithms for OMDb}
\label{OMDb-wordpair-generation-table}
\centering
\begin{tabular}{|c|c|c|c|c|}
\hline 
\multicolumn{1}{|c|}{\bf mAP }  &
\multicolumn{1}{c|}{\bf Baseline} &
\multicolumn{1}{c|}{\bf Semantic} &
\multicolumn{1}{c|}{\bf N-gram} &
\multicolumn{1}{c|}{\bf Non-K} 
\\ \hline 
mAP 1      &0.14134       &\textbf{0.14870}     &0.13202    &0.14302    \\
\hline 
mAP 3      &0.09212       &\textbf{0.09657}     &0.08801    &0.09406    \\
\hline 
mAP 5      &0.07312       &\textbf{0.07635}     &0.07111    &0.07575    \\
\hline 
mAP 10     &0.05113       &\textbf{0.05501}     &0.05132    &0.05585    \\
\hline 
\end{tabular}
\vspace{-1em}
\end{table}

The first thing we observe from the Table~\ref{OMDb-wordpair-generation-table} is that both ``semantic'' word pair generation and ``Non-K'' word pair generation give us better \begin{math} mAP \end{math} score than the baseline; however, the \begin{math} mAP \end{math} score of the ``semantic'' generation is slightly higher than the ``Non-K'' generation. This is because, although both ``Non-K'' and ``semantic'' techniques extract word pairs using natural language processing, without the K-means clustering, semantically similar pairs will be considered separately. Hence there will be lots of redundancies in the input space. This will either increase the size of the input space, or, in order to control the input size, reduce the amount of information captured by the input set. The K-means clustering performs the function of compression and feature extraction.  

The second thing that we observe is that, for the ``N-gram'' word pair generation, its \begin{math} mAP \end{math} score is even lower than the baseline. Beside the OMDb dataset, other two datasets show the same pattern. This is because the ``semantic'' model extracts word pairs from natural language processing, therefore those word pairs have the semantic meanings and grammatical dependencies. However, the ``N-gram'' word pair generation simply extracts words that are adjacent to each other. When introducing some meaningful word pairs, it also introduces more meaningless word pairs at the same time. These meaningless word pairs act as noises in the input. Hence, including word pairs without semantic importance does not help to improve the model accuracy.

\section{Conclusion}\label{sec:conclusion}
In this paper, we proposed a few techniques to preprocess the dataset and optimize the original RBM model. During the dataset preprocessing, first, we used a semantic dependency parser to extract the word pairs from each sentence in the text document. Then, by applying a two-way TF-IDF processing, we filtered the data in word level and word pair level. Finally, K-means clustering algorithm helped us merge the similar word pairs and remove the noise from the feature dictionary. We replaced the original word only RBM model by introducing word pairs. At the end, we showed that proper selection of K value and word pair generation techniques can significantly improve the topic prediction accuracy and the document retrieval performance. With our improvement, experimental results have verified that, compared to original word only RBM model, our proposed word/word pair combined model can improve the \begin{math} mAP \end{math} score up to 10.48\% in OMDb dataset,  up to 1.11\% in Reuters dataset and up to 12.99\% in the 20NewsGroup dataset.






%




\bibliographystyle{IEEEtran}
\bibliography{IEEEabrv,reference}

\end{document}